\pdfoutput=1

\documentclass[11pt]{article}

\usepackage[]{acl}

\usepackage{times}
\usepackage{latexsym}

\usepackage[T1]{fontenc}

\usepackage[utf8]{inputenc}

\usepackage{microtype}
\usepackage{booktabs}
\usepackage{graphicx}
\usepackage{makecell}
\usepackage{xspace}
\usepackage{multirow}
\usepackage{amsmath}
\usepackage{amsthm}
\usepackage{hyperref}
\usepackage{cleveref}
\usepackage{color,colortbl}
\usepackage{subcaption}
\usepackage{amssymb}
\usepackage{pifont}
\usepackage{float}

\newcommand{\cmark}{\ding{51}}%
\newcommand{\xmark}{\ding{55}}%

\newcommand{\rsec}{Section~}
\DeclareFontFamily{T1}{cmtt}{\hyphenchar \font=45}
\newcommand{\method}{\textsc{PGRA}\xspace}

\title{Prompt-Guided Retrieval Augmentation for \\ Non-Knowledge-Intensive Tasks}
\author{
Zhicheng Guo$\textsuperscript{\rm $1,3$}$\Thanks{\ \ E-mail: \href{guo-zc21@mails.tsinghua.edu.cn}{\texttt{guo-zc21@mails.tsinghua.edu.cn}}},
Sijie Cheng$\textsuperscript{\rm $1,2,3,5$}$, 
Yile Wang$\textsuperscript{\rm $2$}$, 
Peng Li$\textsuperscript{\rm $2,4$}$\Thanks{\ \ Corresponding authors. 
E-mail: \href{lipeng@air.tsinghua.edu.cn}{\texttt{lipeng@air.tsin- ghua.edu.cn}}, 
\href{liuyang2011@tsinghua.edu.cn}{\texttt{liuyang2011@tsinghua.edu.cn}}.}\ ,
 \and 
 Yang Liu$^{1,2,3,4}$\footnotemark[2]
\\
 $\textsuperscript{\rm $1$}$Dept. of Comp. Sci. \& Tech., Institute for AI, Tsinghua University, Beijing, China \\
 $\textsuperscript{\rm $2$}$Institute for AI Industry Research (AIR), Tsinghua University, Beijing, China \\
 $\textsuperscript{\rm $3$}$Beijing National Research Center for Information Science and Technology, Beijing, China \\
 $\textsuperscript{\rm $4$}$Shanghai Artificial Intelligence Laboratory, Shanghai, China \\
 $\textsuperscript{\rm $5$}$School of Computer Science, Fudan University, Shanghai, China \\
}

\begin{document}
\maketitle
\begin{abstract}
Retrieval-augmented methods have received increasing attention to support downstream tasks by leveraging useful information from external resources. 
Recent studies mainly focus on exploring retrieval to solve knowledge-intensive (KI) tasks.
However, the potential of retrieval for most non-knowledge-intensive (NKI) tasks remains under-explored. 
There are two main challenges to leveraging retrieval-augmented methods for NKI tasks: 1) the demand for diverse relevance score functions and 2) the dilemma between training cost and task performance.
To address these challenges, we propose a two-stage framework for NKI tasks, named \method.
In the first stage, we adopt a \textit{task-agnostic retriever} to build a shared static index and select candidate evidence efficiently.
In the second stage, we design a \textit{prompt-guided reranker} to rerank the nearest evidence according to task-specific relevance for the reader.
Experimental results show that \method outperforms other state-of-the-art retrieval-augmented methods.
Our analyses further investigate the influence factors to model performance and demonstrate the generality of \method. {Codes are available at \small{\texttt{https://github.com/THUNLP-MT/PGRA}}}. 
\end{abstract}

\section{Introduction}

Retrieval-augmented methods aim at enhancing dense models with non-parametric indices to better leverage external knowledge~\cite{pmlr-v162-borgeaud22a,izacard2022few,wang2022zemi}. 
By decoupling knowledge storage from model parameters, retrieval-augmented methods can achieve comparable or better performance than large-scale pre-trained models with orders of magnitude less parameters on tasks such as language modeling~\cite{khandelwal20generalization, pmlr-v119-guu20a, pmlr-v162-borgeaud22a} and question answering~\cite{lee-etal-2019-latent,karpukhin-etal-2020-dense,izacard-grave-2021-leveraging}. 
Moreover, as external knowledge is stored in the non-parametric index, knowledge can be updated simply by replacing the index without further training~\cite{izacard2022few}. Therefore, retrieval-augmented methods have attracted increasing interest in recent years and achieved promising results in various natural language processing tasks~\cite{zhang-etal-2018-guiding,khandelwal20generalization, pmlr-v119-guu20a,karpukhin-etal-2020-dense}.


Despite their success, retrieval-augmented methods for {the majority} of non-knowledge-intensive (NKI) tasks remain under-explored. 
Following~\citet{NEURIPS2020_6b493230}, we define tasks that ``humans could not reasonably be expected to perform without access to an external knowledge source'' as knowledge-intensive (KI) tasks and the others as NKI tasks. 
Previous studies~\citep{karpukhin-etal-2020-dense,izacard-grave-2021-leveraging,izacard2022few} have extensively explored the potential of retrieval-augmented methods for various KI tasks.
As for NKI tasks, most efforts are devoted to language modeling~\cite{khandelwal20generalization, pmlr-v119-guu20a}, text generation~\cite{NEURIPS2020_6b493230}, and machine translation~\cite{zhang-etal-2018-guiding,khandelwal2021nearest},  although there is a wide range of NKI tasks, such as sentiment analysis~\citep{10.1145/1341531.1341561cr, socher-etal-2013-recursive}, text classification~\citep{hovy-etal-2001-toward, li-roth-2002-learning} and linguistic acceptability~\citep{warstadt2018neural}.
Therefore, we ask this question: \textit{Can retrieval-augmented methods assist on a wider range of NKI tasks?}

\begin{figure*}[t!]
    \centering
    \includegraphics[width=\linewidth]{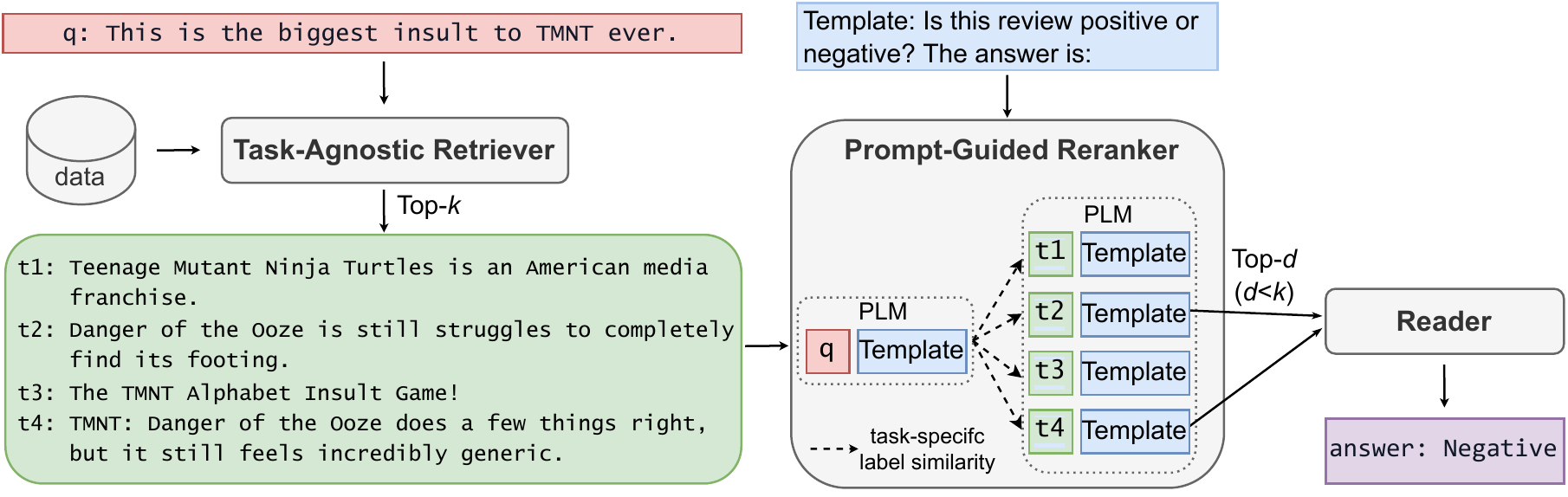}
    \caption{The framework of our proposed Prompt-Guided Retrieval Augmentation (\method) method. We first retrieve candidates through a task-agnostic retriever (\rsec\ref{sec21}), then use a task-specific prompt and pre-trained language model (PLM) to rerank the candidates (\rsec\ref{sec22}). We send the top results to the reader to make predictions. (\rsec\ref{sec23}).}
    \label{fig:framework}
\end{figure*}

However, leveraging retrieval-augmented methods for more types of NKI tasks faces two major challenges.
On the one hand, there is \textit{a demand for diverse relevance score functions}.
To retrieve the most desirable {evidence} from the index, proper relevance score functions are needed. Although the relevance score functions suitable for predicting the next token distribution are well-studied in works on language modeling, text generation, and machine translation, NKI tasks require more diverse  relevance score functions. For example, the text classification task may favor evidence with similar sentence-level semantics~\cite{reimers-gurevych-2019-sentence,gao2021simcse} while the linguistic acceptability task may prefer linguistically similar evidence~\cite{warstadt2018neural}. Therefore, it is non-trivial to satisfy all these diverse requirements in a single framework.
On the other hand, there is \textit{a dilemma between training cost and task performance}. The external knowledge index and retriever are crucial for the performance of a retrieval-augmented method~\cite{lee-etal-2019-latent,karpukhin-etal-2020-dense,izacard-grave-2021-leveraging}. Previous works show that joint training index with the dense model results in better performance~\cite{pmlr-v119-guu20a,xiong2020approximate}. However, due to the large size of external knowledge, updating the index periodically during training is computationally expensive. On the contrary, keeping the index static is computationally cheap but makes it hard to meet the diverse requirements of NKI tasks. Therefore, it is difficult to balance the trade-off between training cost and task performance.

To address these challenges, we propose a two-stage framework, entitled \method, to better retrieve task-specific resources for NKI tasks. The overall framework is shown in Figure~\ref{fig:framework}. In the first stage, we use a task-agnostic retriever to recall candidate {evidence}, which builds a shared static index for all tasks. In the second stage, we adopt prompt-guided pretrained language models (PLMs;~\citealp{brown2020language,zhang2022opt}) as a reranker to rerank the candidates according to the task-specific relevance score functions.
Finally, we feed the reranked top {evidence} to the reader to generate answers. 
By leveraging textual prompts, our framework can satisfy the demand for diverse relevance score functions.
As both the retriever and the reranker are training-free, the expensive computational cost of periodical index update in training is avoided.
At the same time, experimental results justify the effectiveness of our framework on various datasets. Therefore, we successfully break the dilemma between training cost and task performance.

Our main contributions are three-fold: 
\begin{itemize}
    \setlength{\itemsep}{0pt}
    \setlength{\parsep}{0pt}
    \setlength{\parskip}{0pt}
    \item We propose a prompt-guided retrieval augmentation method for a wider range of non-knowledge-intensive tasks, which are hardly explored in previous works.
    \item By combining the retrieval-and-rerank procedure with textual prompts, our framework maintains {reasonably low training cost} while satisfying diverse task-specific relevance score function requirements.
    \item Extensive experimental results and analysis show that our framework is effective for diverse non-knowledge-intensive tasks.
\end{itemize}


\section{Methods}
In this section, we introduce our proposed \underline{P}rompt-\underline{G}uided \underline{R}etrieval \underline{A}ugmentation (\method) method, as shown in Figure~\ref{fig:framework}. 
Our proposed method mainly has three components: (i) a \textit{task-agnostic retriever} using a shared retriever to build static indexes to select top-$k$ candidate evidence from large-scale external resources; 
(ii) a \textit{prompt-guided reranker} adopting PLMs to measure task-specific relevance for reranking candidate evidence; (iii) a \textit{reader} taking the final top-$d$ ($d<k$) ranked evidence as augmentations to generate answers. 

\subsection{Task-Agnostic Retriever}\label{sec21}

Given that the external resource is extremely large-scale, from millions~\cite{khandelwal20generalization, khandelwal2021nearest} to billions~\cite{wang2022zemi, izacard-grave-2021-leveraging, chen-etal-2017-reading}, we use a shared retriever to build the static index once. The key and value of the index are the task-agnostic text representation and the text itself, respectively. The index will be shared across tasks, and thus we save a significant amount of training cost (See \rsec\ref{trainingcost} for discussion). 
Formally, given the input as query $q$, and the external resource containing a bunch of text $ \mathcal{R} = \{t_1,t_2,\cdots,t_{\left|\mathcal{R}\right|}\}$, we firstly encode representations for both query and text, which can be denoted as $\text{Enc}(q)$ and $\text{Enc}(t_i)$, respectively. The representations of text then serve as keys of the index. Then, we use a dense inner product to compute the similarity $\text{Sim}(q, t_i)$ based on the index:
\begin{equation}
\begin{aligned}
    \text{Sim}^\text{agnostic}(q, t_i) = \frac{\exp(\text{Enc}(q)\cdot  \text{Enc}(t_i))}{\sum_{j=1}^{\left|\mathcal{R}\right|}\exp(\text{Enc}(q)\cdot  \text{Enc}(t_j))}.
\end{aligned}
\end{equation}

With the similarity scores, we get the top-$k$ nearest evidence according to retrieval distribution which is the softmax over these scores.
Then we follow the \texttt{faiss}~\cite{johnson2019billion} implementation to efficiently complete the approximate retrieval via Maximum Inner Product Search (MIPS).
These top-$k$ pieces of evidence are regarded as candidates for the second stage for further reranking.

\subsection{Prompt-Guided Reranker}\label{sec22}
As discussed above, the task-agnostic retriever in the first stage selects the nearest candidates by evaluating the similarity of the static indexes between input and external text. 
However, such shared retrievers neglect the fact that different NKI tasks prefer their own task-specific relevance score functions, which is crucial to retrieve useful evidence.

In order to meet the demand for diverse relevance score functions, we further design a task-specific reranker in the second stage.
To avoid expensive calculations for training a task-specific retriever per NKI task, we exploit the in-context learning ability of prompt-guided PLMs.

\begin{table}[t!]
    \centering
    \resizebox{\columnwidth}{!}{
    \begin{tabular}{l}
    \toprule
    \rowcolor[gray]{0.95} 
    \textbf{Template: SST-2} \\
    \midrule
    \textcolor{gray}{/* Example */} \\
    \makecell[l{p{8cm}}]{Does the following sentence have a positive or negative sentiment?  \\
     one long string of cliches .\\
    The answer is negative. \\
    \\
    \textcolor{gray}{/* Test data */} \\
    Does the following sentence have a positive or negative sentiment?  \\
    the performances take the movie to a higher level .\\
    The answer is }
    \\
    \bottomrule
    \end{tabular}}
    \caption{The prompt instances of in-context learning in our prompt-guided retriever. We use 8 examples per prompt. More details can be found in Appendix \ref{app:prompts}.}
    \label{tab:incontext}
\end{table}

At first, we adopt in-context learning under the few-shot setups to encode task-specific representations of the input query $q$ and the top-$k$ pieces of evidence $\{e_1,e_2,\cdots,e_k\}$.
Specifically, we design a prompt specialized for each task by concatenating $m$ exemplars randomly sampled from the training datasets with manually written task descriptions as shown in Table~\ref{tab:incontext}.
Then, we feed an auto-regressive PLM (\textit{e.g.}, OPT;~\citealp{zhang2022opt}) with both constructed prompts and our input to obtain the task-specific representations of the next predicted tokens:
\begin{equation}
\begin{aligned}
{\rm prefix} &= p_1,l_1,p_2,l_2,\cdots,p_m,l_m \\
h_q^* &= {\rm PLM} ( [\![ {\rm prefix};p_q ]\!] )\\
h_{e_i}^* &= {\rm PLM} ( [\![ {\rm prefix};p_{e_i} ]\!] ),
\end{aligned}
\end{equation}
where $p_1, p_2, \cdots, p_m$ are the $m$ prompts of the examplars, $l_1, l_2, \cdots, l_m$ are the labels, $p_q$ and $p_{e_i}$ ($i =1,\cdots,k$) are the prompts of the input query and evidence $e_i$, respectively. 
The prefix text is then concatenated to the prompts of the query or the evidence as the textual input.
Lastly, the inputs are fed to the model to generate the last hidden states of the first new token $h_q^*\in \mathbb{R}^d$ and $h_{e_i}^*\in \mathbb{R}^d$.

It is worth noting that text in the external knowledge resource may lack explicit labels for NKI tasks.
Through in-context learning with prompt guidance, the representations of the inputs and external evidence encoded by the PLM implicitly contain different critical features to solve various tasks.
Similar to the first stage, we compute the similarity between the representations of input $q$ and its candidate evidence $e_i$, which reflects their task-specific relevance:
\begin{equation}
\begin{aligned}
\text{Sim}^\text{task-specific}(q,e_i) = \frac{\exp(h_{q}^* \cdot h_{e_i}^*)}{\sum_{j=1}^{k}{\exp(h_{q}^* \cdot h_{e_j}^*})}.
\end{aligned}
\end{equation}
Finally, we rerank the candidate evidence according to the aforementioned task-specific relevance score and select the top-$d$ results for the reader in the next section.

\begin{table*}[t!]
    \centering
    \small
    \resizebox{\textwidth}{!}{
    \begin{tabular}{lcccccccccc}
    \toprule
        \textbf{Method} & \textbf{Retrieval} & \textbf{SST-2} & \textbf{SST-5} & \textbf{CoLA} & \textbf{TREC} & \textbf{CR} & \textbf{MR} & \textbf{MPQA} & \textbf{Subj}  &\textbf{Average} \\
        \midrule
        {\textbf{ICL} (OPT-13b)} & \xmark & 93.0 & 46.0 & 1.8 & 26.8 & 73.2 & 61.7 & 71.6 & 51.1  & 53.2\\
        \midrule
        \rowcolor[gray]{0.95} 
        \multicolumn{11}{c}{\textbf{\textit{T5-base} (220M)}}\\
        {\textbf{$k$-NN}} & \cmark & 59.2 & 22.8 & 1.0 & 28.0 & 51.8 & 55.1 & 52.6 & 72.1 & 42.9\\
        {\textbf{LM-BFF}} & \xmark & 86.0 & 45.5 & 5.5 & 76.2 & 90.0 & 83.1 & 82.3 & 90.2 & 69.9\\
        {\textbf{T5-base}} & \xmark & 91.3 & {56.7} & 30.4 & 80.4 & 89.8 & 89.4 & 89.2 & 96.0 & 77.9\\
        {\textbf{RAG}} & \cmark & \underline{93.0} & \textbf{57.5} & {\textbf{58.5}} & 80.4 & 87.2 & \underline{90.2} & 89.5 & 96.5  & 81.6\\
        {\textbf{FiD}} & \cmark & 92.2 & 56.6 & 56.9 & \underline{ 80.8 }& \underline{91.3} & 90.1 & \underline{89.8} & \underline{96.6} & \underline{81.8}\\
        {\textbf{\method} (Ours)} & \cmark &\textbf{93.9}  & \underline{56.9} & \underline{57.0} & \textbf{80.8} & \textbf{91.7} & \textbf{91.1} & \textbf{90.3} & \textbf{97.0} &   \textbf{82.3} \\

        \midrule 
        \rowcolor[gray]{0.95} 
        \multicolumn{11}{c}{\textbf{\textit{T5-large} (770M)}}\\
        \textbf{$k$-NN} & \cmark & 64.5& 23.6& 2.1& 28.8 & 56.8 &58.2&53.7 & 72.4 & 45.0\\
        \textbf{LM-BFF} & \xmark & 90.8 & 49.0 & 6.9 & 70.6 & 91.1 & 83.5 & 89.5 & 88.4 & 71.2\\
        {\textbf{T5-large}} & \xmark & \underline{95.2} & 59.2 & \underline{60.7} & \underline{80.8} & 92.1 & 91.5 & \textbf{90.7} & 97.3  & 83.4\\
        {\textbf{RAG}} & \cmark & \underline{95.2} & 57.2 & 60.1 & 80.2 & 91.2 & 92.1 & \underline{90.6} & 96.4 &  82.9\\
        {\textbf{FiD}} & \cmark & 94.8 & \underline{59.5} & 60.2 & \underline{80.8} & \underline{92.4} & \textbf{92.5} & \underline{90.6} & \textbf{97.5} & \underline{83.5}\\
        {\textbf{\method} (Ours)} & \cmark & \textbf{95.7} & \textbf{59.8} & \textbf{61.1} & \textbf{80.9} & \textbf{92.6} & \underline{92.4} & \underline{90.6} & \textbf{97.5} & \textbf{83.8} \\
    \bottomrule
    \end{tabular}
    }
    \caption{{The results of baselines and our \method. For models with T5-base backbone, we use $d=16$. For models with T5-large backbone, we use $d=8$ in the second stage due to GPU memory limitation. The best results are \textbf{bolded}, and the second-best ones are \underline{underlined}.}}
    \label{tab:main_table1}
\end{table*}



\subsection{Reader}\label{sec23}
To encode useful information from the reranked evidence and infer the final answer for the query text $q$, we use the FiD (Fusion-in-Decoder;~\citealp{izacard-grave-2021-leveraging}) model as our reader, which has a Seq2seq pre-trained Transformer~\cite{NIPS2017_3f5ee243} such as T5~\cite{2020t5}. 
Specifically, each piece of evidence obtained from the reranker is concatenated with the query, which is independently fed into the encoder. The decoder takes the embeddings of these concatenations produced by the encoder and computes cross attention over them to give the final answer prediction. Following the prompt-based learning~\cite{schick-schutze-2021-exploiting,liu2021pre}, we transfer the NKI tasks to the form of language modeling, where the answers are deduced according to the label prediction in a context. The overall reader is trainable and the parameters are updated given the training samples of the required NKI tasks.




\section{Experiments}

\subsection{Experimental Setups}
\label{sec:exp_setup}
\paragraph{Tasks and Metrics.} Following the setups in LM-BFF~\citep{gao-etal-2021-making}, We conduct the experiments mainly on four types of NKI tasks: (1) Sentiment analysis. We use a various of datasets from different domains, including SST-2~\cite{socher-etal-2013-recursive} and SST-5~\cite{socher-etal-2013-recursive} for the general domain with two and five labels, CR~\cite{10.1145/1341531.1341561cr} for comment reviews, MR~\cite{pang-lee-2004-sentimental} for movie reviews, MPQA~\cite{Wiebe2005AnnotatingEO} for news opinions;
(2) Linguistic acceptability. We adopt CoLA~\cite{warstadt2018neural}, which aims to discriminate whether a sentence is grammatically correct; 
(3) Question classification. We use TREC~\cite{hovy-etal-2001-toward,li-roth-2002-learning}, in which a question needs to be classified into six categories;
(4) Subjectivity analysis. We use Subj~\cite{pang-lee-2004-sentimental}, which has to judge whether the sentence is subjective or objective.
As for metrics, we report Matthew's correlation for CoLA while reporting accuracy in all other tasks.
More details about datasets and metrics can be found in Appendix~\ref{app:datasets}.

\paragraph{External Resources and Models.} 
As for the external resources, we use Wiki1M following~\citet{gao2021simcse}. 
Furthermore, in the first stage, we use BERT-base-uncased~\cite{devlin-etal-2019-bert} as our shared task-agnostic retriever. We also compare with other retrievers of the first stage in \Cref{sec:retrievers}. 
In the second stage, we use OPT-13b~\cite{zhang2022opt} as our auto-regressive PLMs to obtain the task-specific representations. We further explore the effects on the size of our PLMs in \rsec\ref{sec:opt_size}. 
Finally, we adopt T5-base and T5-large~\cite{2020t5} as our readers to generate answers. 

\paragraph{Implementation Details.} 
We use the checkpoints of T5-base, T5-large, and OPT-13b from HuggingFace\footnote{\url{https://huggingface.co/models}}. Our manually designed prompts are obtained from PromptSource~\cite{bach2022promptsource}. 
We finetune the T5 model on each task with the AdamW~\cite{loshchilov2017decoupled} optimizer. We search hyper-parameters of learning rate of 
\{1e-5, 2e-5, 5e-5, 8e-5, 1e-4\} and batch sizes of \{4, 8\}. 
We set the number of top-$k$ in the first stage to 150, while the number of top-$d$ in the second stage is 16 with T5-base and 8 with T5-large due to computational resource limitation.
We further compare the effect of $k$ and $d$ in \rsec\ref{sec:kd}. 
We use 8 shots for prompts during reranking in the second stage.
Our experiments are conducted with one NVIDIA V100 GPU.

\paragraph{Baselines.}
We compare our proposed method \method with the following baselines:
(1) In-context learning (ICL;~\citealp{brown2020language}), which directly uses OPT-13b, the same as our PLM in the second stage, to generate answers under the few-shot setups ({8 shots in our settings});
(2) T5~\cite{2020t5}, which use T5-base and T5-large in supervised learning;
(3) $k$-Nearest Neighbour ($k$-NN;~\citealp{DBLP:journals/corr/abs-2004-04523}), in which the model makes a majority vote based on distances between embeddings;
(4) LM-BFF~\cite{gao-etal-2021-making}, which is a few-shot inference method tuned with dedicated prompts;
(5) RAG~\cite{NEURIPS2020_6b493230}, which treats context samples as hidden variables and jointly trains the retriever and generator; 
(6) FiD~\cite{izacard-grave-2021-leveraging}, which concatenates query and context samples in the encoder and generates answers with cross attention.
To ensure a fair comparison, we uniformly adopt the same reader (\textit{i.e.}, T5-base and T5-large) for retrieval-augmented methods. As for $k$-NN and LM-BFF, we also use T5-base and T5-large for building representations and training.
In the baseline of in-context learning, we use the same templates as ours in the second stage.


\subsection{Results}
We compare our proposed \method with the aforementioned baseline methods, where the results are shown in \Cref{tab:main_table1}. We include results on both T5-base and T5-large models for generality reasons. {We run our experiments three times and report details of each run in Appendix~\ref{sec:multiple_runs}. We report average results here and first-run results in the analysis section below.}  

Firstly, the \method can significantly outperform the simple $k$-Nearest Neighbour and few-shot methods, including in-context learning with OPT-13b and LM-BFF. As for the $k$-Nearest Neighbour, it is simply based on the distances of embeddings encoded by T5. As for the few-shot methods, in-context learning uses prompts to elicit PLMs to generate answers without updating parameters. It is worth noting that we use in-context learning with OPT-13b as our prompt-guided reranker in the second stage. The performance of in-context learning is ordinary, so it is surprising that it can assist on \method. We will further discuss the reason behind this in \rsec\ref{sec:label_consistency}. 
Meanwhile, LM-BFF is further fine-tuned on the prompts to give answers. Thus, its performance is obviously higher than $k$-Nearest Neighbour and in-context learning with OPT-13b but remains a large gap to \method.


Secondly, compared to supervised learning (\textit{i.e.}, T5-base and T5-large) and retrieval-augmented baselines, \method still outperforms them across most tasks. Specifically, the line of retrieval methods with a T5-base reader outperforms supervised learning with the T5-base model, while retrieval-augmented methods with a T5-large reader are worse or comparable to supervised learning with the T5-large model. 
Furthermore, our method \method can obviously surpass these baselines, in both T5-base and T5-large setups.
In conclusion, extensive experimental results have shown that our \method
is effective on diverse NKI tasks.







\section{Analysis}

\subsection{Effects of Label Consistency}
\label{sec:label_consistency}
In this section, we probe the influence of retrieved evidence on the model performance of our \method from the aspect of label consistency.
Note that our external text is without any task-specific labels.
Therefore, we use a T5-base model fine-tuned on the specific task, which is the closest to our \method reader but without retrieval, to generate \textit{pseudo-label} for all text in the external resource. 
In detail, if the pseudo-label of evidence is the same as the ground-truth label of the input, we say the evidence is \textit{consistent} with the input.
We can then directly detect the relation between the number of consistent evidence and model performance  at the instance level.
Specifically, out of 16 pieces of total retrieved evidence, the number of consistent evidence with the same (pseudo) labels as the input varies from 0 and 16.

\begin{figure}[]
    \centering  
     \begin{subfigure}[b]{\linewidth}
        \includegraphics[width=\columnwidth]{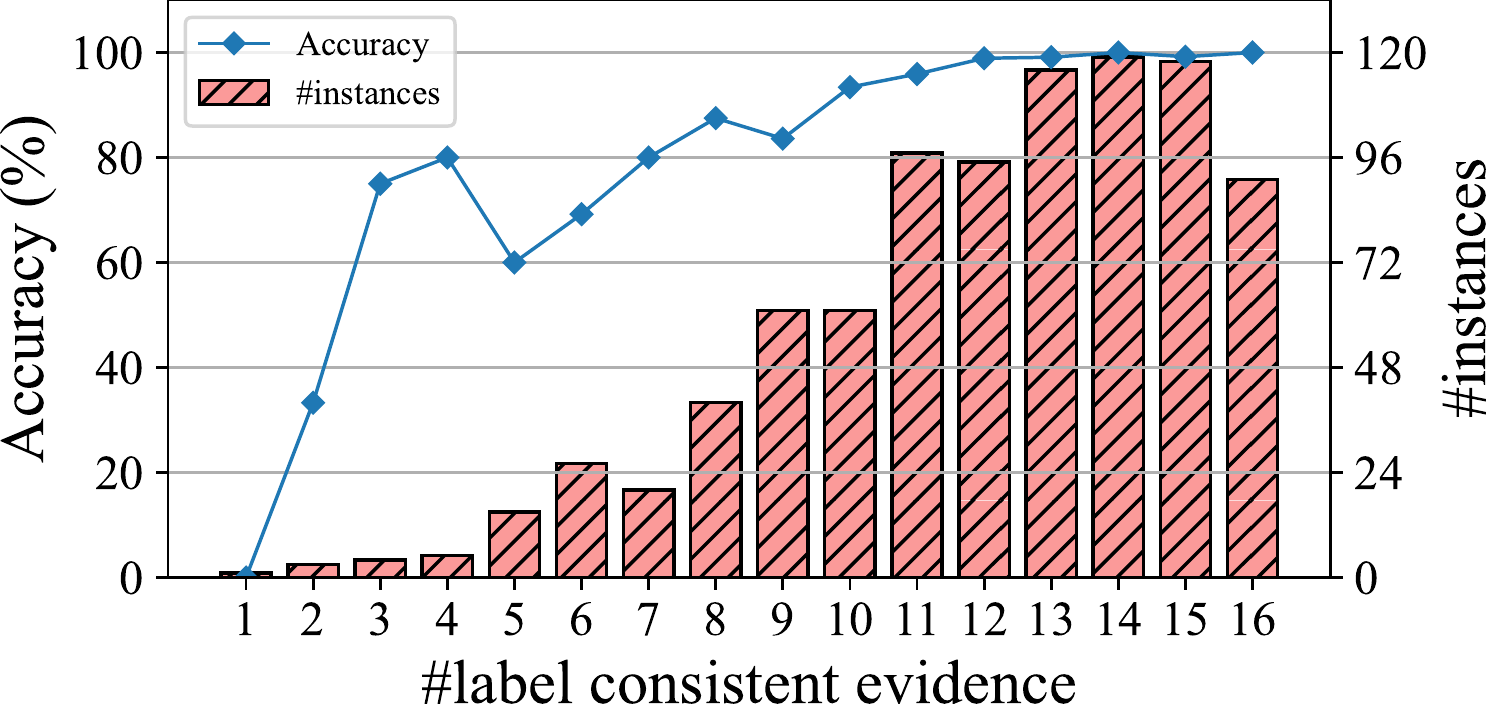}
        \caption{\method}
        \label{fig:label_ours}
      \end{subfigure}
      \hfill
      \begin{subfigure}[b]{\linewidth}
        \includegraphics[width=\columnwidth]{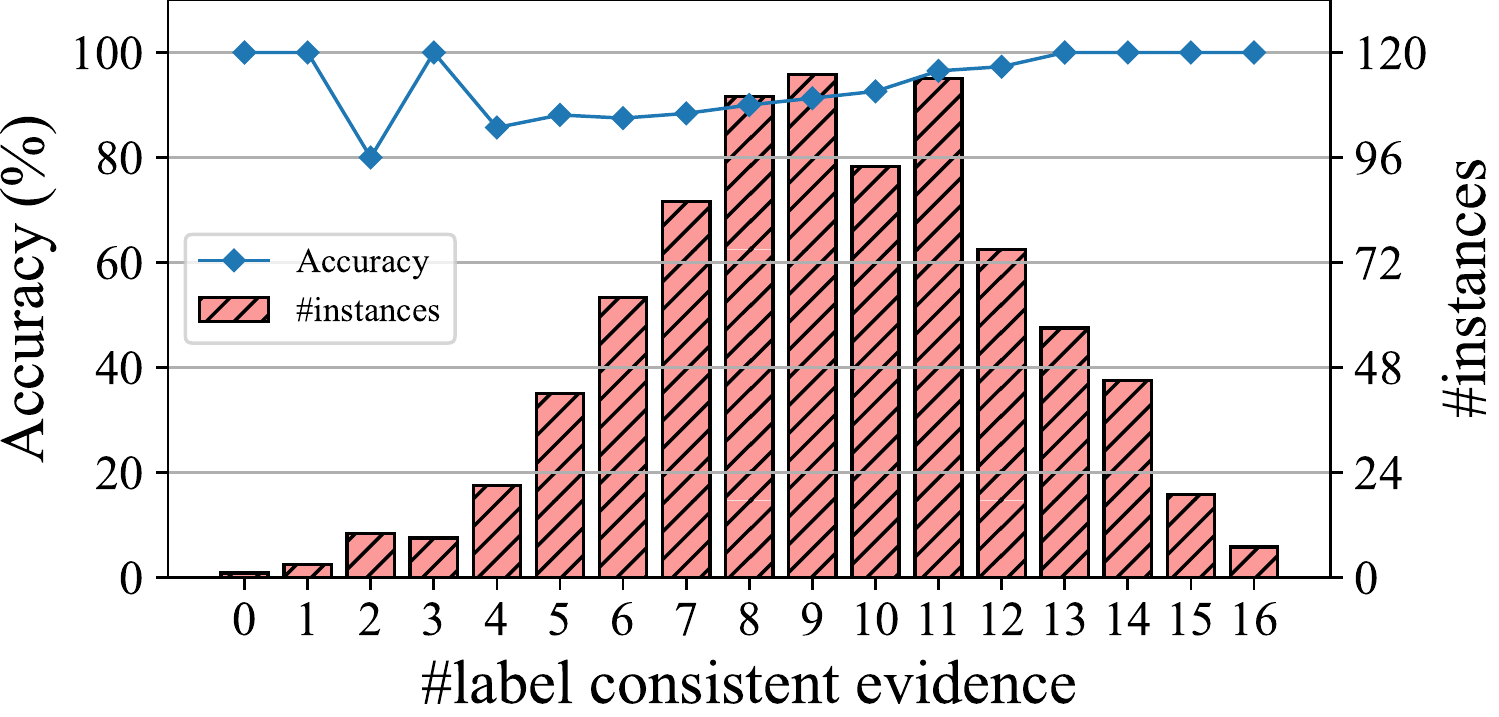}
        \caption{FiD}
        \label{fig:label_fid}
      \end{subfigure}
    
    \caption{{The pseudo label consistency of samples in SST-2 with \method and FiD (T5-base models for both). We plot the accuracy scores of instances with different numbers of label-consistent evidence, along with the number of such instances. }}
    \label{fig:labelsst2}
\end{figure}

Taking the SST-2 task as an example,
we count the total number of instances with different numbers of consistent evidence. 
We then compute the average accuracy of \method for the instances with the same number of consistent evidence. The results are shown in \Cref{fig:label_ours}.
Firstly, since we rerank the evidence based on the relevance score of pseudo-labels, the number of instances also rises as the number of consistent evidence increases.
The phenomenon indicates that we can always find sufficient task-specific evidence retrieved from the first stage, except for a small part of inputs which is possibly caused by the limitation of the $k$'s size in the first stage. 
Secondly, the average accuracy is also rising as the number of consistent evidence increases, which reflects that the model performance is related to the (pseudo) label consistency. However, when the number of consistent evidence is small (\textit{i.e.}, 3 and 4), the accuracy can also be high. This is because the number of instances is too small, so the result is insignificant. Furthermore, it is interesting to find that when the number of consistent evidence is high enough (\textit{i.e.}, larger than 13), the accuracy approaches 100\%, which shows that there exists high potential in increasing label consistency to improve model performance.

{We use the same method to plot the label consistency figure on the FiD baseline, shown in \Cref{fig:label_fid}. As can be seen from the figure, it still holds that the more label-consistent evidence, the higher accuracy the model can achieve. The difference between \method and FiD is that \method can retrieve more label-consistent evidence than FiD. }

\subsection{Effects of $k$ and $d$}
\label{sec:kd}
\begin{figure}[t!]
    \centering
    \includegraphics[width=\columnwidth]{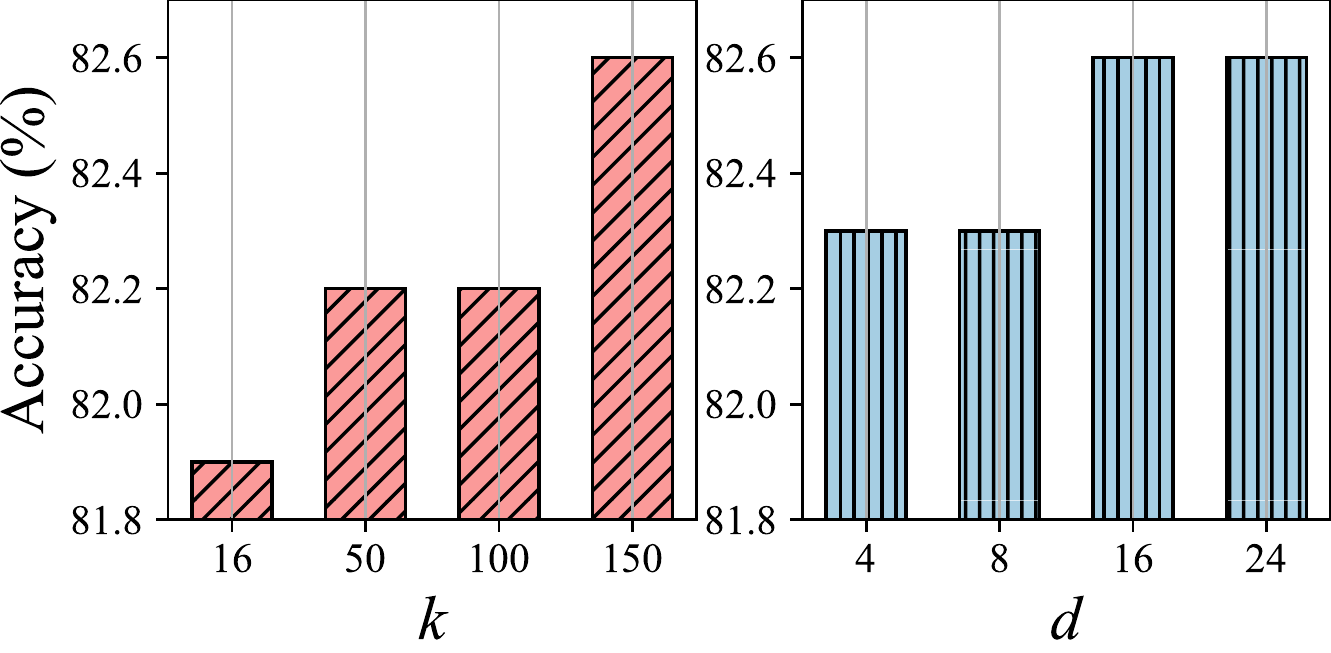}
\caption{Accuracy against $k$ (left) and $d$ (right). Details of performance on different tasks can be found in \Cref{app:kdsize}.}
\label{fig:top_kd}
\end{figure}

In this section, we further investigate the effects of $k$ and $d$ on the performance, where $k$ and $d$ are the numbers of final retrieved evidence in the first and second stages, respectively.
In detail, we run \method with different $k$ or different $d$, while other setups keep the same as main experiments.
As seen from \Cref{fig:top_kd}, larger $k$ values can consistently improve the average performance, while larger $d$ values maintain a relatively stable trend.
As for $k$, larger $k$ values mean providing more candidate evidence for the second stage reranker to find more appropriate instances with (pseudo) label consistency.
As for $d$, larger $d$ values indicate more consistent evidence if the proportion of consistent evidence keeps the same. 
At the same time, their top consistent evidence is the same, and the candidate evidence is fixed with the same $k$, so their performance is close. 
In our expectation, the \method can better solve diverse NKI tasks with larger $k$ if enough computing resources are allowed.

\subsection{Effects of OPT Model Sizes}
\label{sec:opt_size}

\begin{figure}[]
    \centering
    \includegraphics[width=\columnwidth]{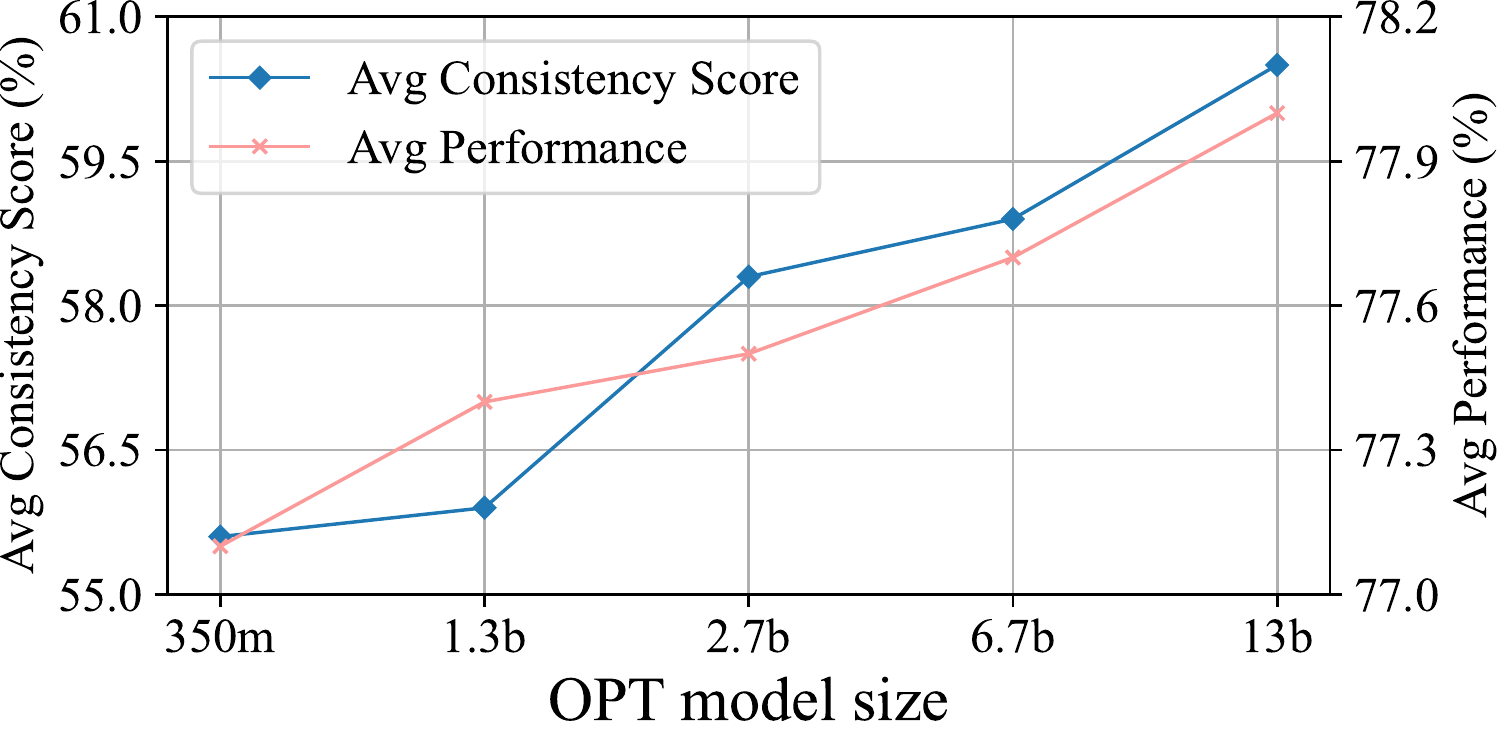}
    \caption{Average performance and average consistency score on 5 tasks (SST-2, SST-5, CoLA, MR and MPQA) against different OPT model sizes. Detailed information can be found in Appendix \ref{app:kdsize}. }
    \label{fig:label_opt}
\end{figure}

In this section, we first investigate the effects on the performance of different sizes of the OPT models used in the prompt-guided reranker. Specifically, we vary the size of OPT models and conduct experiments in five downstream tasks. The model performances are shown in the orange line of \Cref{fig:label_opt}.
The overall trend is obviously that the larger OPT models can achieve better performance.
We believe that the larger OPT models  have better abilities to apply task-specific features to encode representations, and further obtain more effective task-specific relevance scores to retrieve evidence.


To validate this assumption, we further investigate the relations between (pseudo) label consistency and model performance of different OPT model sizes. We define the pseudo-label consistency score (\textit{i.e.}, \textit{consistency score}) as the proportion of retrieved instances with the same pseudo-label as the input. 
For example, given the input with a positive ground-truth label, when our \method recalls 5 consistent and 3 inconsistent pieces of evidence, the consistency score is 5/8 $=$ 62.5\%. 
As shown in the \Cref{fig:label_opt}, overall, larger models with higher consistency scores result in better performance, which is within expectation.


\subsection{Effects of Evidence Granularity}
\begin{figure}[htbp!]
    \centering
    \includegraphics[scale=0.24]{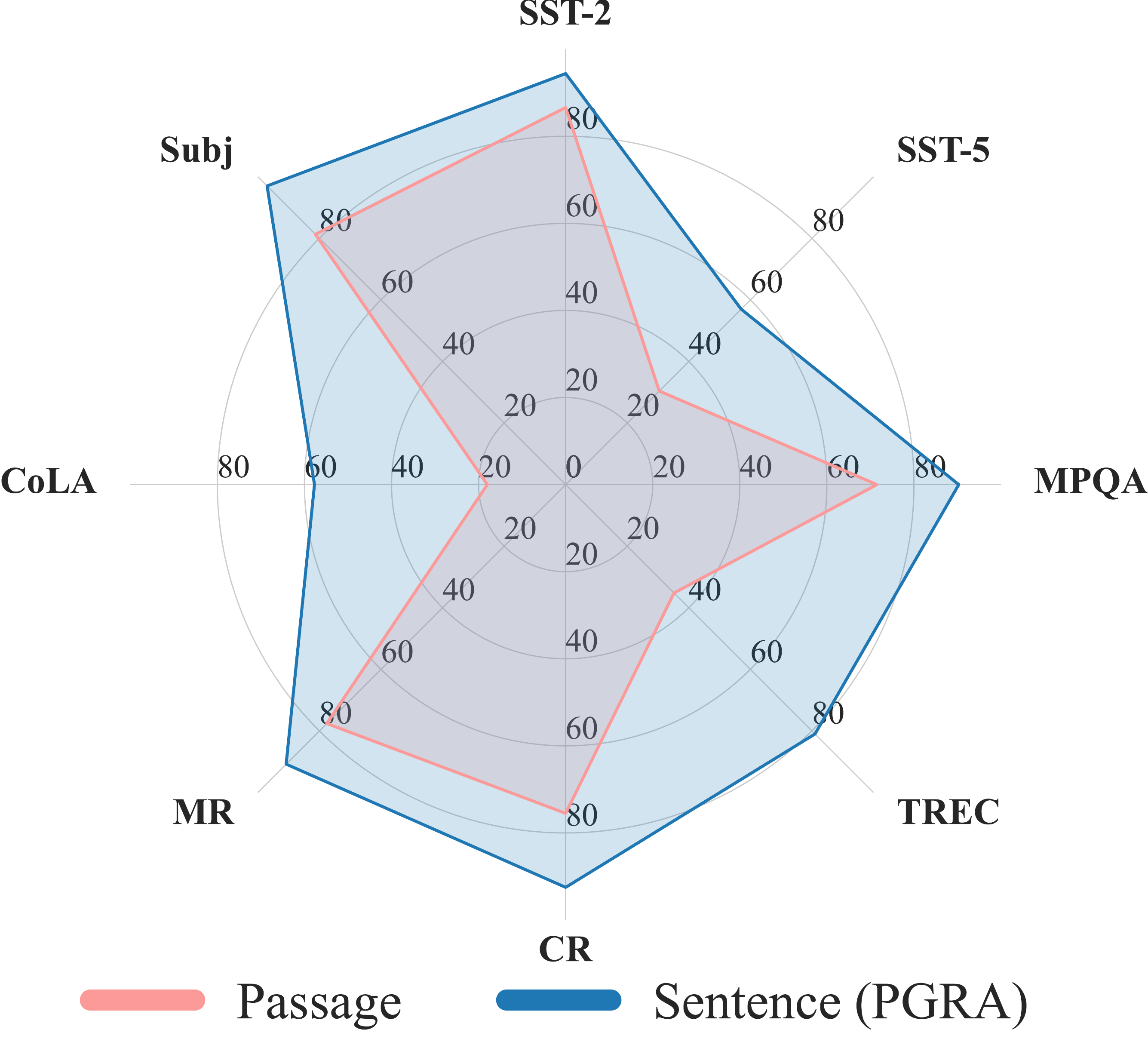}
    \caption{Performance of \method with passage-level and sentence-level external datastores.}
    \label{fig:passage}
\end{figure}

In this work, we propose to use a task-specific relevance score to retrieve from sentence-level external resources, rather than popular passage-level used in previous studies~\citep{chen-etal-2017-reading, izacard-grave-2021-leveraging, pmlr-v119-guu20a}.
To demonstrate that our granularity of external evidence is appropriate, we compare the model performance between sentence-level and passage-level evidence.
As for passage-level evidence, we use WikiDPR~\citep{chen-etal-2017-reading} as external resources. We randomly sample 1M passages from WikiDPR to keep the same data size as our sentence-level external resource in the main experiment.
The results are shown in Figure~\ref{fig:passage}. Across all NKI tasks, our sentence-level setup performances significantly surpass passage-level setup.
This phenomenon indicates that sentence-level evidence can better satisfies the task-specific demands for NKI tasks. For example, it is easier to show a clear sentiment orientation in a sentence than in a paragraph.

\subsection{Training Cost}\label{trainingcost}

To solve the dilemma between training cost and task performance, we propose \method where both the retriever and reranker are training-free.
To demonstrate this statement, in this section, we approximately compare the training cost of our method \method with training a task-specific retriever per task, the latter of which needs periodical refreshing indexes (\textit{i.e.}, refreshed-index models).
{Considering a significant amount of training time concentrates on building and refreshing indexes}, we mainly statistic this part.
Due to the limitation of computation resources, we conduct our main experiment on 1M data from Wikipedia.
In our \method, we only need to build the index once without extra training, and the time cost $c$ is about 0.5 hours. However, although the time cost $c$ of building index is almost the same, they need to periodically refresh the index $n$ times to learn a task-specific retriever.
Lastly, for all $h$ tasks, their total training cost is $c \times n \times h$, which is much larger than our time cost $c$. It is worth noting that the external resource is usually much larger than ours~\citep{chen-etal-2017-reading, wang2022zemi, izacard2022few}, so the gap between refreshed-index models and our \method will further grow to explode.

        

 \begin{figure}[t!]
     \centering
     \includegraphics[scale=0.55]{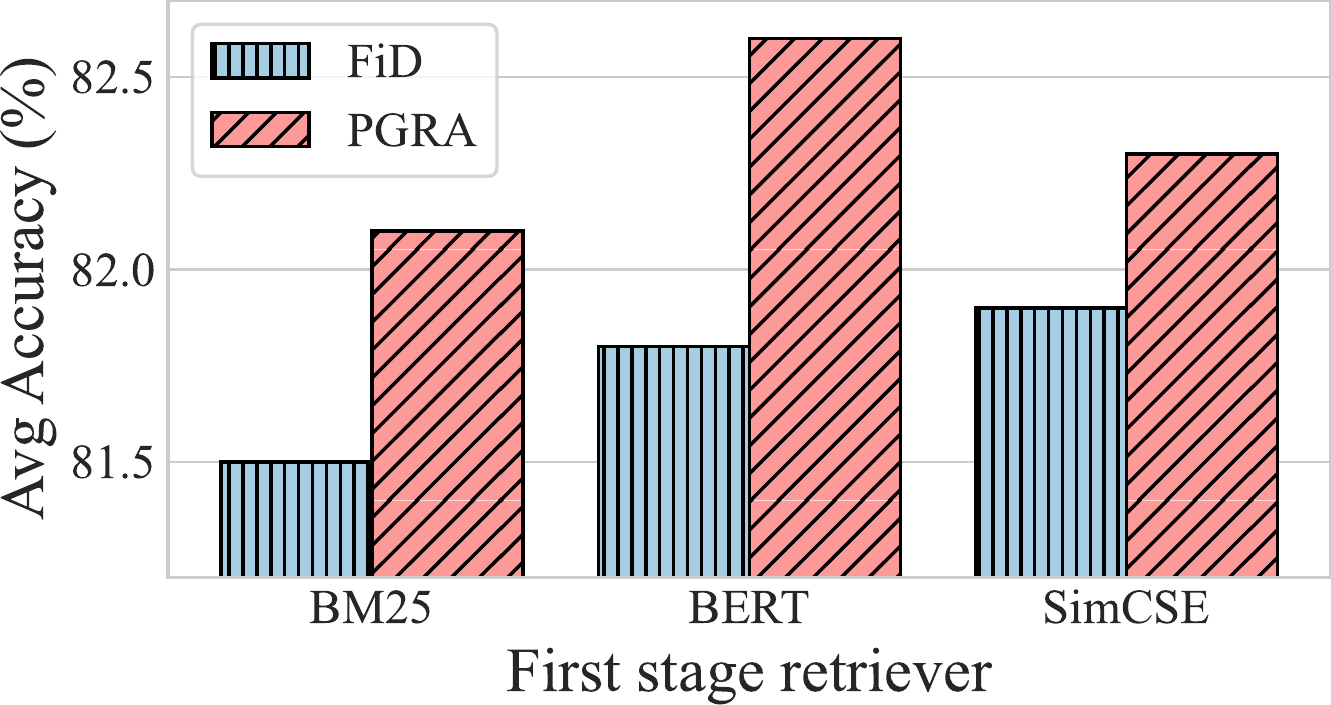}
         
     \caption{Comparison between FiD and our \method with BM25, BERT and SimCSE retrievers. More details of specific performance in all tasks can be found in Appendix~\ref{app:retrievers}.}
     \label{fig:generality}
 \end{figure}

\subsection{Generalization on Retrievers}
\label{sec:retrievers}
In this section, we study the generalization of \method with different first-stage retrievers. We use popular retrievers like BM25~\cite{bm25},  BERT~\cite{devlin-etal-2019-bert}, SimCSE~\cite{gao2021simcse} to compare FiD and our \method. As shown in \Cref{fig:generality}, our method \method consistently outperforms FiD, no matter which retriever to use.
This phenomenon indicates that \method can adapt to different types of retrievers in the first stage to solve various NKI tasks.
Furthermore, the retriever with BM25 performs worse than both BERT and SimCSE counterparts, which is consistent with previous studies~\citep{sparse-dense-survey}. 

\subsection{Case Study}
In \Cref{tab:case}, we present a case in SST-2 with different retrieved evidence from baselines (\textit{i.e.}, FiD and RAG) and our \method. As shown in the table, our \method can exactly predict the correct answer, while both FiD and RAG are wrong. 
To further analyze the retrieved evidence from different methods, we find that sentences retrieved by FiD and RAG may have overlapped tokens or similar semantics.
For example, the retrieved evidence from FiD is highly related to filming and stories, consistent with the ``title'', ``characters'', and ``camera'' in the input.
But their retrieved evidence hardly has the same sentiment orientation to assist the downstream task.
Some of them may have even opposite sentiments, such as the second sentence retrieved by FiD.
However, our retrieved evidence from \method clearly has a negative sentiment orientation, though some may not have explicitly relatedness with the query, such as the second retrieved sentence. 
In general, evidence retrieved by our \method method based on task-specific relevance can effectively improve performance on NKI tasks.

\begin{table}[t!]
    \small
    \centering
    \resizebox{\linewidth}{!}{
    \begin{tabular}{p{0.15\columnwidth}p{0.85\columnwidth}}
    \toprule
    \rowcolor[gray]{0.95} \multicolumn{2}{c}{{\textbf{Dataset: SST-2}} }\\
    Input & The title not only describes its main characters  but the lazy people behind the camera as well. \\
    Label & \textit{Negative} \\
    \midrule
    \rowcolor[gray]{0.95} \multicolumn{2}{c}{{\textbf{Method: FiD}} }\\
    Prediction & \textit{Positive} \\
    Evidence & {(1) The story overlaps science fiction, theology, and philosophy.}\\
     & (2) However, the film’s greatness is not limited to a few isolated scenes. \\
    \midrule
    \rowcolor[gray]{0.95} \multicolumn{2}{c}{{\textbf{Method: RAG}} }\\
    Prediction & \textit{Positive} \\
    Evidence & {(1) The 1978 King Cup was the 20th season of the knockout competition since its establishment in 1956.} \\ 
     & (2) Per Kristian Norvik was born in Vadsø, Norway on February 10, 1938. \\
    \midrule
    \rowcolor[gray]{0.95} \multicolumn{2}{c}{{\textbf{Method: \method}} }\\
    {Prediction} & \textit{Negative} \\
    Evidence & {(1) Once it had been shown that the film could not be realized, ``The Works'' was officially abandoned.}\\
    & (2) The play can also be seen as a discussion of romanticism and reality, in a quite disillusional way. \\     
    \bottomrule
    \end{tabular}}
    \caption{Case study of FiD, RAG, and our \method with the top-2 retrieved evidence in SST-2.}
    \label{tab:case}
\end{table}

\section{Related Work}
{Retrieval-augmented methods} are widely used for knowledge-intensive tasks such as question answering~\cite{chen-etal-2017-reading,karpukhin-etal-2020-dense,izacard-grave-2021-leveraging,izacard2022few}, where explicit knowledge is required to achieve reasonable performance, even for human~\cite{NEURIPS2020_6b493230}. Such systems usually follow a \textit{retriever-reader} architecture, where an existing retriever like BM25~\cite{bm25} or a trained dual-encoder~\cite{lee-etal-2019-latent,luan-etal-2021-sparse} is used, followed by a reader model to fuse the retrieved results. We focus on non-knowledge-intensive tasks and 
propose a prompt-guided retrieval method for mining fine-grained textual information across multiple tasks, without training a specific retriever for each of them. Recently, \citet{wang2022zemi} also applied retrieval-augmented methods to more general tasks by keeping a shared BM25 retriever unchanged for each task while modifying the reader for information filtering. In contrast, we propose a two-stage retrieval method to find task-specific information at a low cost for different downstream tasks.

{Prompt-based methods} gained much advance in recent years~\cite{schick-schutze-2021-exploiting,liu2021pre,gao-etal-2021-making}, where downstream tasks can be solved via transforming the problem to the form of language modelling. Combined with PLMs such as GPT-3~\cite{brown2020language} and OPT~\cite{zhang2022opt}, such methods show strong performance under zero-shot or few-shot settings. Recently, there are also some works that leverage prompts for retrieval. For example, \citet{asai2022task} collected large-scale instruction-annotated datasets for training instruction-guided retrievers for tasks, \citet{van2022don} use prompts for searching regex-based patterns from unlabeled corpora. 
Our method is inspired by these works and different in that we leverage the pre-trained models for retrieving according to task-specific relevance and propose an efficient retrieval-augmented method for NKI tasks.

\section{Conclusion}
In this paper, considering the demand for diverse relevance score functions to solve wider NKI tasks, we propose a two-stage method \method.
In the first stage, we use a task-agnostic retriever for building shared static indexes to select candidate evidence. In the second stage, we design a prompt-guided reranker to rerank candidates with task-specific relevance for the reader. 
Extensive experimental results show that our proposed method \method can overall outperform previous state-of-the-art retrieval-augmented methods. Furthermore, we explore the influence of label consistency between input and retrieved evidence from the prompt-guided reranker and demonstrate the generality of our \method on both evidence granularities and types of retrievers.
In the future, we will consider ways to improve the pseudo-label consistency to enhance model performances according to our analyses.

\section*{Limitations}
In this work, we present \method to retrieve task-specific context evidence to support NKI tasks. However, our work has some limitations. Firstly, we have not experimented with our \method on sentence-pair tasks, such as MRPC~\cite{dolan-brockett-2005-automatically}, in which the model needs to infer the relationship between two sentences. Retrieving two sentences from an external datastore is non-trivial as there are hardly sentence pairs in the Wikipedia datastore. A larger corpus with more diverse data sources may help in this case. Secondly, 
We restrict our \method to classification tasks but not generation tasks. Similar to sentence-pair tasks, retrieving sentences that may help the model generate text is more complex. For example, data related to both the source and the target may help in machine translation~\cite{khandelwal2021nearest}. We will research this question in the future.
{Last but not least, we have not extensively tested the performance of our method on KI tasks, except for some preliminary analysis in \Cref{app:ki}.} This restricts the generality of our methods. Solving KI tasks depends on knowledge in the passage-level external datastore while matching such information needs possibly more specialized prompts for our method. Thus, it is for our future work.

\section*{Ethics Statement}
Currently, large language models with retrieval augmentation require a large amount of computation in indexing a large-scale datastore, retrieving from that large datastore and refreshing index during training. Despite improving model performance, the retrieval augmentation methods need too much computation power. This not only limits the usability of such models but also harms the fairness in this community. Our work tries to balance the performance of retrieval augmentation methods and the training cost, in that our method does not need to retrain a new retriever and rebuild an index when facing a new task. This may help the community in developing new low-cost methods.

During selecting the external datastore and tasks, we follow previous studies and choose the well-known Wikipedia dataset and common tasks. Biases from the data may be reflected in the results. In addition, when using the model on a larger scale, more consideration needs to be paid to deal with biases in retrieved text.

\section*{Acknowledgements}
This work is supported by the National Key R\&D Program of China (2022ZD0160502) and the National Natural Science Foundation of China (No. 61925601, 62276152, 62236011). We also want to show our gratitude to Ziyue Wang for her suggestions on the writing of the paper.

\bibliography{acl_zhicheng_sim,custom}

\appendix


\clearpage
\section{Multiple Runs of the Main Experiment}
\label{sec:multiple_runs}
{We run \method three times under the settings of the main experiment in \Cref{tab:main_table1} and report results of these runs in \Cref{tab:multiple_runs}.}

\section{Experiments with More Retrieved Evidence For FiD baselines}

\label{app:fid_more}
{We run additional experiments for FiD (T5-base) baseline with more retrieved evidence. The results are shown in \Cref{tab:fid_more}. It can be seen that with more retrieved evidence, although the average scores become higher, FiD still underperforms \method.}
\begin{table}[htbp!]
    \centering
    \small
    \begin{tabular}{lcccc}
    \toprule
        \textbf{Datasets} & $d=8$ & $d=48$ \\
        \midrule
        SST-2 & 92.2 & 93.3 \\
        SST-5 & 56.6 & 56.8 \\
        CoLA & 56.9 & 56.3 \\
        TREC & 80.8 & 81.0 \\
        CR & 91.3 & 91.2 \\
        MR & 90.1 & 90.8 \\
        MPQA & 89.8 & 89.5 \\
        Subj & 96.6 & 96.8\\
        \midrule
        Avg. & 81.8 &  82.0 \\
    \bottomrule
    \end{tabular}
    \caption{{Detailed analysis of the impact of top-$d$ in the second stage.}}
    \label{tab:fid_more}
\end{table}

\section{Impact of $k,d$ and OPT Model Sizes}
\label{app:kdsize}
We explore the impact of $k,d$ and second-stage OPT model sizes. The full analysis is shown in \rsec\ref{sec:kd} and  \rsec\ref{sec:opt_size}. \Cref{tab:top_k}, \Cref{tab:top_d} and \Cref{tab:opt_size} show detailed performance of our method in each task. For each ablation, we keep other hyper-parameters the same as used in \Cref{tab:main_table1}.

\begin{table}[htbp!]
    \centering
    \small
    \resizebox{\columnwidth}{!}{
    \begin{tabular}{lcccc}
    \toprule
        \textbf{Datasets} & $k=16$ & $k=50$ & $k=100$ & $k=150$ \\
        \midrule
        SST-2 & 93.6& 92.8 & 93.7 & 94.4 \\
        SST-5 & 55.7 & 56.4 & 56.0 & 57.0\\
        CoLA & 56.0 & 58.9 & 56.8 & 57.7 \\
        TREC & 80.6 & 80.6 & 80.6 & 81.0 \\
        CR & 91.7 & 91.8 & 92.1 & 92.5  \\
        MR & 90.8 & 90.5 & 91.5 & 90.8 \\
        MPQA & 90.1 & 90.2 & 89.8 & 90.3 \\
        Subj & 97.0 & 96.7 & 97.0 & 97.0 \\
        \midrule
        Avg. & 81.9 & 82.2 & 82.2 & 82.6\\
    \bottomrule
    \end{tabular}}
    \caption{Detailed analysis of the impact of top-$k$ in the first stage.}
    \label{tab:top_k}
\end{table}

\section{Label Consistency}
\label{app:label_consistency}
We include the details of label consistency scores of our \method with different second-stage OPT models on each task in \Cref{tab:stage2_label}.

\begin{table}[htbp!]
    \centering
    \small
    \begin{tabular}{lcccc}
    \toprule
        \textbf{Datasets} & $d=4$ & $d=8$ & $d=16$ & $d=24$ \\
        \midrule
        SST-2 & 94.4 & 94.3  & 94.4 & 94.2 \\
        SST-5 & 56.2 & 56.5  & 57.0 & 57.8\\
        CoLA & 57.3 & 55.7  & 57.7 & 56.4 \\
        TREC & 80.8 & 80.6  & 81.0 & 80.8 \\
        CR & 91.7 & 92.2  & 92.5 & 92.1 \\
        MR & 90.9  & 91.5 & 90.8 & 91.8 \\
        MPQA & 90.1  & 90.4 & 90.3 & 90.5 \\
        Subj & 96.7  & 97.0 & 97.0 & 97.0\\
        \midrule
        Avg. & 82.3 & 82.3 & 82.6 & 82.6\\
    \bottomrule
    \end{tabular}
    \caption{{Detailed analysis of the impact of top-$d$ in the second stage.}}
    \label{tab:top_d}
\end{table}

\begin{table}[htbp!]
    \centering
    \small
    \begin{tabular}{lccccccccc}
    \toprule
        \textbf{Datasets} & learning rate & batch size\\
        \midrule
        {SST-2} &  8e-5 & 8 \\
        {SST-5} & 2e-5 & 8\\
        {CoLA} & 8e-5  & 8 \\
        {TREC} & 8e-5 & 4 \\
        {CR} & 1e-5 & 8\\
        {MR} & 1e-4 & 8\\
        {MPQA} & 8e-5 & 4 \\
        {Subj} & 8e-5 & 8 \\
    \bottomrule
    \end{tabular}
    \caption{Information of the tasks.}
    \label{tab:besthyper}
\end{table}

\begin{table*}[t!]
    \centering
    \small
    \begin{tabular}{lccccccccc}
    \toprule
        \textbf{Runs}  & \textbf{SST-2} & \textbf{SST-5} & \textbf{CoLA} & \textbf{TREC} & \textbf{CR} & \textbf{MR} & \textbf{MPQA} & \textbf{Subj}  &\textbf{Average} \\
        \midrule
        \rowcolor[gray]{0.95} 
        \multicolumn{10}{c}{\textbf{\textit{T5-base} (220M)}}\\

    Run 1 &  {94.4}  & {57.0} & {57.7} & {81.0} & {92.5} & {90.8} & {90.3} & {97.0} & {82.6} \\        
    Run 2 & 93.5 & 56.5 & 57.6 & 80.8 & 90.6 & 91.3 & 90.1 & 97.0 & 82.2 \\
    Run 3 & 93.9 & 57.1 & 57.5 & 80.6 & 91.9 & 91.2 & 90.4 & 97.0 & 82.5 \\
    Average & 93.9 & 56.9 & 57.6 & 80.8 & 91.7 & 91.1 & 90.3 & 97.0 & 82.4 \\
    Std & 0.45 & 0.32 & 0.10 & 0.20 & 0.97 & 0.26 & 0.15 & 0.0 & 0.21 \\

        \midrule 
        \rowcolor[gray]{0.95} 
        \multicolumn{10}{c}{\textbf{\textit{T5-large} (770M)}}\\

       Run 1 & 96.0  & {59.4} & {64.0} & {81.2} & {92.8} & {93.0} & {90.8} & {97.6} & {84.4} \\
       Run 2 & 95.4 & 60.1 & 60.1 & 81.0 & 92.0 & 92.3 & 90.5 & 97.3 & 83.6 \\
       Run 3 & 95.8 & 60.0 & 59.1 & 80.6 & 92.9 & 91.9 & 90.5 & 97.6 & 83.6 \\
       Average & 95.7 & 59.8 & 61.1 & 80.9 & 92.6 & 92.4 & 90.6 & 97.5 & 83.8 \\
       Std & 0.31 & 0.38 & 2.59 & 0.31 & 0.49 & 0.56 & 0.17 & 0.17 & 0.45 \\
    \bottomrule
    \end{tabular}
    \caption{Multiple run results of \method.}
    \label{tab:multiple_runs}
\end{table*}

\begin{table*}[p]
    \centering
    \small
    \begin{tabular}{lccccccccc}
    \toprule

        \textbf{Datasets} & \textbf{SST-2} & \textbf{SST-5} & \textbf{CoLA} & \textbf{TREC} & \textbf{CR} & \textbf{MR} & \textbf{MPQA} & \textbf{Subj} & \textbf{Average} \\
        \midrule
        OPT 350m &  93.9 & 56.2 & 55.1 & 80.8 & 90.8 & 90.3 & 90.0 & 97.2 & {81.8} \\
        OPT 1.3b & 93.8 & 57.8 & 54.8 & 80.6 & 92.4 & 90.5 & 90.0 & 96.7 & 82.1 \\
        OPT 2.7b & 94.2 & 56.5 & 55.4 & 81.0 & 91.6 & 90.7 & 90.9 & 96.7 & 82.1 \\
        OPT 6.7b & 93.6 & 57.5 & 56.2 & 80.6  & 92.2 & 90.6 & 90.6 & 96.8 & 82.3 \\
        OPT 13b & 94.4 & 57.0 & 57.7 & 81.0 & 92.5 & 90.8 & 90.3 & 97.0 & 82.6\\
    \bottomrule
    \end{tabular}
    \caption{Detailed analysis of the impact of OPT sizes with $k=150,d=16$.}
    \label{tab:opt_size}
\end{table*}

\begin{table*}[t!]
    \centering
    \small
    \begin{tabular}{lccccccc}
    \toprule
        \textbf{Encoder} & \textbf{SST-2} & \textbf{SST-5} & \textbf{CoLA} & \textbf{MR} & \textbf{MPQA} & \textbf{Average} & \textbf{Performance} \\

        \midrule
        OPT 350m & 63.4 & 28.8 & 69.0 & 60.3 & 56.6 & 55.6 & 77.1 \\
        OPT 1.3b & 68.3 & 32.2 & 69.3 & 55.6 & 53.6 & 55.9 & 77.4\\
        OPT 2.7b & 68.9 & 32.5 & 69.0 & 65.8 & 55.2 & 58.3 & 77.5\\
        OPT 6.7b & 70.1 & 32.0 & 69.1 & 69.3 & 54.2 & 58.9 & 77.7 \\
        OPT 13b & 75.2 & 30.2 & 69.1 & 70.5 & 57.3 & 60.5 & 78.0 \\

    \bottomrule
    \end{tabular}
    \caption{Pseudo-label consistency with different OPT models. ``Average'' is the average label consistency score on the five tasks.  Performance is the average of the 5 tasks. We keep $k=150, d=16$ in this experiment.}
    \label{tab:stage2_label}
\end{table*}

\section{Generality on Retrievers}
\label{app:retrievers}
We include the detailed performance of FiD and our \method on all tasks with different first-stage encoders, namely BM25, BERT and SimCSE. The results are shown in \Cref{tab:generality}.

\section{Generalization on KI tasks}
\label{app:ki}
{We perform experiments on the FEVER~\cite{thorne-etal-2018-fever} benchmark.  FEVER is a fact verification task, requiring a model to classify whether a claim is factually correct. Due to resource limitations, we sample 5k claim-label pairs from the training set and 1k pairs from the validation set. We run FiD and \method with both T5-base backbone and keep other hyperparameters the same as in \Cref{tab:main_table1}. Note that we did this experiment with a sentence-level datastore (Wiki1m). FiD and \method achieve 73.8\% and 77.7\% accuracy respectively. The results confirm again the performance increase with \method. However, one might notice that the performance of FiD with a traditional passage-level datastore can achieve better performance. We acknowledge this as a limitation of our method because a passage-level datastore requires much different relevance metrics as stated in the Limitation section. This is also a possible future direction.}

\begin{table*}[t!]
    \centering
    \small
    \begin{tabular}{lccccccccc}
    \toprule
        \textbf{Method} & \textbf{SST-2} & \textbf{SST-5} & \textbf{CoLA} & \textbf{TREC} & \textbf{CR} & \textbf{MR} & \textbf{MPQA} & \textbf{Subj} & \textbf{Average} \\
        \midrule
        FiD (BM25) & 93.6 & 56.4 & 56.7 & 80.6 & 90.2 & 89.7 & 88.2 & 96.8 & 81.5 \\
        \method (BM25) & 93.9 & 56.6 & 56.9 & 80.4 & 91.4 & 90.4 & 90.5 & 96.5 & 82.1\\
        \midrule
        FiD (BERT) & 92.2 & 56.6 & 56.9 & { 80.8 }& {91.3} & 90.1 & {89.8} & {96.6} & {81.8} \\
        \method (BERT) & 94.4  & 57.0 & 57.7 & 81.0 & 92.5 & 90.8 & {90.3} & {97.0} & {82.6} \\
        \midrule 
        FiD (SimCSE) & 93.2 & 56.8 & 56.2 & 81.0 & 90.9 & 90.5 & 90.1 & 96.3 & 81.9 \\
        \method (SimCSE) & 93.5 & 57.7 & {55.9} & 81.0 & 92.0 & 91.0 & 90.4 & 96.9 & 82.3\\

    \bottomrule
    \end{tabular}
    \caption{Table of generalization performance of FiD and \method with different first-stage encoders.}
    \label{tab:generality}
\end{table*}

\section{Datasets and Metrics}
\label{app:datasets}
We use the Wiki1M from SimCSE~\cite{gao2021simcse} as our external datastore. This dataset is a subset of Wikipedia and used in \cite{gao2021simcse}. We report information on tasks in \Cref{tab:dataset_info}. We use the same configuration as \cite{gao-etal-2021-making}, including dataset splits.
\begin{table*}[hbt!]
    \centering
    \small
    \begin{tabular}{lccccccccc}
    \toprule
        \textbf{Datasets} & Type  & Labels & Avg Input length\\
        \midrule
        \textbf{SST-2} & Sentiment analysis &  positive, negative & 19\\
        \textbf{SST-5} & Sentiment analysis &  v. pos., positive, neutral, negative, v. neg.& 18\\
        \textbf{CoLA} & Linguistic acceptability & acceptable, unacceptable & 8\\
        \textbf{TREC} & Question classification & abbr., entity, description, human, loc., num. & 10\\
        \textbf{CR} & Sentiment analysis &  positive, negative & 19\\
        \textbf{MR} & Sentiment analysis &  positive, negative & 20\\
        \textbf{MPQA} & Sentiment analysis &  positive, negative& 3\\
        \textbf{Subj} & Subjectivity analysis & subjective, objective & 23\\
    \bottomrule
    \end{tabular}
    \caption{Information of the tasks and datasets.}
    \label{tab:dataset_info}
\end{table*}

\section{Training Details}
As stated in \Cref{sec:exp_setup}, we search hyper-parameters of learning rate of \{1e-5, 2e-5, 5e-5, 8e-5, 1e-4\} and batch sizes of \{4, 8\}. We train our models for 5000 steps on the training set. The best hyperparamters found are shown in \Cref{tab:besthyper}.

\section{Prompts}
We include all prompts used in all 8 tasks in Table~\ref{tab:app_incontext}.
\label{app:prompts}
\begin{table*}[t!]
    \small
    \centering
    \resizebox{\textwidth}{!}{
    \begin{tabular}{p{\textwidth}}
    \toprule
    \rowcolor[gray]{0.95} 
    \textbf{Template: SST-2/CR/MR/MPQA} \\
    \midrule
    \textcolor{gray}{/* Example */} \\
    \makecell[l]{Does the following sentence have a positive or negative sentiment?  \\
     one long string of cliches .\\
    The answer is negative. \\
    \\
    \textcolor{gray}{/* Test data */} \\
    Does the following sentence have a positive or negative sentiment?  \\
    the performances take the movie to a higher level .\\
    The answer is } \\
    \midrule
    \rowcolor[gray]{0.95} \textbf{Template: SST-5} \\
    \midrule
    \textcolor{gray}{/* Example */} \\
    \makecell[l]{What sentiment does this sentence have? terrible, bad, okay, good or great
    "with a romantic comedy plotline straight from the \\
    ages, this cinderella story doesn't have a single surprise up its sleeve ." \\
    The answer is bad \\
    \\
    \textcolor{gray}{/* Test data */} \\
    What sentiment does this sentence have? terrible, bad, okay, good or great \\
    hardly a film that comes along every day. \\
    The answer is } \\
    \midrule
    \rowcolor[gray]{0.95} \textbf{Template: CoLA} \\
    \midrule
    \textcolor{gray}{/* Example */} \\
    \makecell[l]{The following sentence is either "acceptable", meaning it is grammatically correct and makes sense, or "unacceptable". Which \\
    is it? \\
    I ordered if John drink his beer. \\
    The answer is unacceptable \\
    \\
    \textcolor{gray}{/* Test data */} \\
    The following sentence is either "acceptable", meaning it is grammatically correct and makes sense, or "unacceptable". Which\\
    is it? \\
    Angela characterized Shelly as a lifesaver.\\ 
    The answer is }\\
    \midrule
    \rowcolor[gray]{0.95} \textbf{Template: Subj} \\
    \midrule
    \textcolor{gray}{/* Example */} \\
    \makecell[l]{Is this a subjective or objective description? \\
    when the skittish emma finds blood on her pillow why does she still stay behind?\\
    The answer is objective \\
    \\
    \textcolor{gray}{/* Test data */} \\
    Is this a subjective or objective description? \\
    "at the end of the worst day of his life, bruce angrily ridicules and rages against god and god responds ."\\
    The answer is }\\
    \midrule
    \rowcolor[gray]{0.95} 
    \textbf{Template: TREC} \\
    \midrule
      \makecell[l]{    \textcolor{gray}{/* Example */} \\
      Which category best describes the following question: \\
     How far is it from Denver to Aspen.\\
     Choose from the following list: Description, Entity, Abbreviation, Person, Quantity, Location. \\
    The answer is Quantity. \\
    \\
    \textcolor{gray}{/* Test data */} \\
    Which category best describes the following question: \\
     What were Ottoman objectives?\\
     Choose from the following list: Description, Entity, Abbreviation, Person, Quantity, Location. \\
    The answer is }  \\
    \bottomrule
    \end{tabular}
    }
    \caption{The prompt instances of in-context learning in our prompt-guided reranker.}
    \label{tab:app_incontext}
\end{table*}


\end{document}